\PassOptionsToPackage{unicode}{hyperref}
\PassOptionsToPackage{hyphens}{url}
\documentclass[11pt,a4paper]{article}
\usepackage[margin=25mm]{geometry}
\usepackage{amsmath,amssymb}
\usepackage{iftex}
\ifPDFTeX
  \usepackage[T1]{fontenc}
  \usepackage[utf8]{inputenc}
  \DeclareUnicodeCharacter{2248}{\ensuremath{\approx}}
  \DeclareUnicodeCharacter{00B1}{\ensuremath{\pm}}
  \DeclareUnicodeCharacter{00D7}{\ensuremath{\times}}
  \DeclareUnicodeCharacter{03B1}{\ensuremath{\alpha}}
  \usepackage{textcomp} 
\else 
  \usepackage{unicode-math} 
  \defaultfontfeatures{Scale=MatchLowercase}
  \defaultfontfeatures[\rmfamily]{Ligatures=TeX,Scale=1}
\fi
\usepackage{lmodern}
\ifPDFTeX\else
\fi
\IfFileExists{upquote.sty}{\usepackage{upquote}}{}
\IfFileExists{microtype.sty}{
  \usepackage[]{microtype}
  \UseMicrotypeSet[protrusion]{basicmath} 
}{}
\makeatletter
\@ifundefined{KOMAClassName}{
  \IfFileExists{parskip.sty}{%
    \usepackage{parskip}
  }{
    \setlength{\parindent}{0pt}
    \setlength{\parskip}{6pt plus 2pt minus 1pt}}
}{
  \KOMAoptions{parskip=half}}
\makeatother
\usepackage{xcolor}
\usepackage{longtable,booktabs,array}
\usepackage{calc} 
\usepackage{etoolbox}
\makeatletter
\patchcmd\longtable{\par}{\if@noskipsec\mbox{}\fi\par}{}{}
\makeatother
\IfFileExists{footnotehyper.sty}{\usepackage{footnotehyper}}{\usepackage{footnote}}
\makesavenoteenv{longtable}
\usepackage{graphicx}
\makeatletter
\def\maxwidth{\ifdim\Gin@nat@width>\linewidth\linewidth\else\Gin@nat@width\fi}
\def\maxheight{\ifdim\Gin@nat@height>\textheight\textheight\else\Gin@nat@height\fi}
\makeatother
\setkeys{Gin}{width=\maxwidth,height=\maxheight,keepaspectratio}
\makeatletter
\def\fps@figure{htbp}
\makeatother
\providecommand{\tightlist}{%
  \setlength{\itemsep}{0pt}\setlength{\parskip}{0pt}}
\ifLuaTeX
  \usepackage{selnolig}  
\fi
\IfFileExists{bookmark.sty}{\usepackage{bookmark}}{\usepackage{hyperref}}
\IfFileExists{xurl.sty}{\usepackage{xurl}}{} 
\hypersetup{
  hidelinks,
  pdfcreator={LaTeX via pandoc}}

\begin{document}

\title{Multimodal Conditioning of Fine-Tuned Stable Diffusion XL for Controllable and Culturally Faithful Ulos Motif Generation}
\author{%
\begin{minipage}{0.92\textwidth}\centering
Humasak Simanjuntak\textsuperscript{1*}, Tamara Yunika Sianipar\textsuperscript{1},\\
Bronson T.M Siallagan\textsuperscript{1}, Difya Laurensya Ambarita\textsuperscript{1},\\
Arlinta Barus\textsuperscript{1}\\[0.5em]
\small \textsuperscript{1}Faculty of Informatics and Electrical Engineering,\\
\small Institut Teknologi Del, Indonesia\\
\small *humasak@del.ac.id
\end{minipage}}
\date{}
\maketitle

\textbf{Abstract. T}he traditional Batak Ulos weaving industry faces growing challenges in producing diverse, innovative motifs due to limitations in conventional, manually driven design methods. This study proposes a multimodal generative framework integrating a fine-tuned Latent Diffusion Model (Stable Diffusion XL v1.0 via LoRA) with a Multimodal Large Language Model (LLaMA 1.5-7B) to enable controllable, culturally faithful Ulos motif generation. Four complementary conditioning mechanisms: text, image, representation, and semantic map (via ControlNet) jointly guide the generation process, each governing a distinct aspect from semantic intent to spatial layout. A five-level ablation study across three scenarios (shape transformation, colour variation, and high-complexity input) shows that conditioning effectiveness is not proportional to the number of mechanisms combined: Text+Image+Semantic Map achieved the best FID (≈270) and CLIP Score (0.65--0.70) but the weakest SSIM (≈0.65), while Text+Image+Representation offered the best overall balance, with stable SSIM (≈0.84) and competitive FID (≈280). Combining all four mechanisms yielded the weakest FID (≈330), indicating conflicting optimization signals. Qualitative evaluation by nine weavers and thirty public participants confirmed statistically significant positive acceptance (Wilcoxon, p = 0.007 and p \textless{} 0.001, respectively). A web-based prototype supporting text-to-image and image-to-image generation was also developed, offering a practical digital design tool for cultural heritage preservation.


\section{Introduction}

Ulos is a traditional weaving textile of the Batak ethnic group in North Sumatra, Indonesia, functioning not merely as a decorative product but as a cultural symbol embodying social values, kinship, and blessings across generations[1][2][3]. Its geometric composition, symbolic ornamentation, and colour arrangement are traditionally inspired by nature and daily life, transformed into repetitive patterns that carry specific philosophical meanings tied to customary ceremonies such as births, weddings, and funerals [4][5][6][7]. Preserving Ulos therefore requires maintaining both its visual identity and its embedded cultural values, not merely its surface appearance.

Despite this significance, Ulos usage is increasingly confined to formal ceremonies, while younger generations favour contemporary clothing, weakening the transmission of traditional weaving knowledge. Diversifying Ulos into modern products (fashion, accessories, interior design) offers a path to sustain interest among younger audiences, but this requires continuous creation of new motif variations that remain recognizably authentic. This is difficult in practice: weavers rely on tacit knowledge accumulated over years, and manual design tends to produce either overly repetitive motifs or modifications that erode cultural identity. Generating motifs that are simultaneously innovative and culturally faithful thus remains an open problem at the intersection of computational creativity and heritage preservation.

Prior computational efforts have targeted this problem with mixed success. The DiTenun platform explored image quilting for texture synthesis [8], SinGAN for pattern generation from a single image [9], and StyleGAN for motif variation [10], but each is limited: image quilting produces repetitive, unnatural transitions on complex structures; SinGAN\textquotesingle s stochastic output is difficult to control and prone to artefacts at larger scales; and StyleGAN suffers from mode collapse, restricting output diversity. More recently, fine-tuned latent diffusion models (Protogen v3.4 and Stable Diffusion v1.4) were shown to generate higher-quality, structurally coherent motifs, though conditioning remained limited to manually authored text-and-image prompts, without mechanisms capturing richer visual-semantic detail [11]. In a related domain, GenBatik combined StyleGAN2-ADA with a Diffusion-GAN mechanism for batik motif synthesis, improving FID/KID over standard StyleGAN, yet precision and recall remained low and generated motifs still closely resembled training data, indicating that hybrid StyleGAN-diffusion approaches only partially resolve the fidelity--diversity trade-off [12], a limitation especially relevant for motifs with strict compositional rules such as Ulos.

To date, no prior study has combined latent diffusion models with richer multimodal conditioning to generate Ulos motifs that are both controllable and culturally authentic. This study addresses that gap using Stable Diffusion XL (SDXL) v1.0 [13][14][15] as the generative backbone, guided by four complementary conditioning mechanisms: text conditioning for high-level semantic intent, image conditioning for structural fidelity to reference motifs, representation conditioning (via a multimodal LLM) for richer latent-level visual description, and semantic map conditioning (via ControlNet) for spatial layout control. Jointly, these mechanisms enable finer-grained control over generation than text prompts alone, while the SDXL backbone is fine-tuned specifically on the Ulos motif domain to keep outputs faithful to authentic Batak weaving patterns.

The primary contribution of this study is a multimodal generative framework that couples the high-fidelity synthesis capability of latent diffusion models with the semantic conditioning strength of multimodal large language models, enabling controllable Ulos motif generation that is directed through combined visual and textual conditioning rather than manually engineered prompts alone, offering a novel solution to the dual challenge of generating motifs that are both innovative and culturally authentic to Batak tradition.

\section{Materials and Method}
\subsection{Research Design}

This study proposes a generative framework for producing Batak Ulos motifs that are both controllable and faithful to cultural values. The framework integrates three core components: (1) Stable Diffusion XL (SDXL) v1.0 [15], a latent diffusion model fine-tuned using Low-Rank Adaptation (LoRA); (2) LLaVA-1.5-7B, a Multimodal Large Language Model employed as a representation conditioning mechanism[16][17]; and (3) ControlNet [18], which utilizes semantic maps as a spatial conditioning mechanism. These components jointly enable four complementary conditioning modalities: text, image, representation, and semantic map, which are progressively integrated to guide the motif generation process while preserving the visual characteristics and cultural conventions of traditional Ulos textiles.

The effectiveness of the proposed framework is evaluated through a five-level ablation study: (1) a baseline SDXL model without fine-tuning, using only text and image conditioning as a reference configuration; (2) an SDXL model fine-tuned with LoRA, using text and image conditioning; (3) the addition of semantic map conditioning via ControlNet; (4) the inclusion of representation conditioning generated by LLaVA; and (5) the complete framework combining all four conditioning modalities. Each ablation level is further evaluated across three generation scenarios: motif and shape transformation, colour variation (recolouring), and high-complexity input, with three strength values (0.65, 0.75, and 0.85) tested for each scenario.

The generated results are assessed using two complementary evaluation approaches. For quantitative evaluation, the Fréchet Inception Distance (FID) is used to measure the statistical distance between the feature distributions of generated and reference images via an Inception Network feature extractor, with lower FID values indicating closer resemblance to the original data distribution. The Structural Similarity Index Measure (SSIM) evaluates the structural similarity between generated and reference images using three perceptual components: luminance, contrast, and structure. In addition, CLIP-Score is employed to assess the model\textquotesingle s accuracy in translating culturally relevant textual instructions, such as traditional colours and distinctive geometric patterns, into visual representations, while also identifying culturally inconsistent patterns or model hallucinations.

For qualitative evaluation, a questionnaire-based assessment is conducted involving 9 traditional Ulos weavers and 30 public participants. The collected responses are analyzed using hierarchical statistical testing procedures to examine perceived visual quality, cultural authenticity, and the overall effectiveness of the proposed framework.

\subsection{Ulos Motif Dataset}

The Ulos motif dataset was collected from two primary sources: the DiTenun platform, providing 231 raw motif images, and field data gathered directly from traditional Ulos weavers in Silaen and the TB Silalahi Center Museum, contributing six additional images to enhance the diversity of motif variations. After quality screening to remove blurred, distorted, low-resolution, or visually inconsistent samples, a curated dataset of 125 representative images was obtained. Each image represents a cropped unit of a complete Ulos motif to preserve structural integrity, and the dataset was categorized by visual scale into small (18 × 18 px), medium (36 × 36 px), and large (72 × 72 px) groups. All images were saved in .jpg and .png formats and resized to 512 × 512 pixels to match the standard input resolution of latent diffusion architectures such as SDXL. Data augmentation via rotation and horizontal/vertical flipping was applied to expand the training set from 125 to 431 images while preserving the semantic and structural characteristics of the motifs.

Among the 125 curated images, seven were purposively selected as the testing set to capture variations in motif shape, pattern, and colour across four Batak sub-ethnic groups. This strategy reflects two considerations: the limited availability of authentic Ulos images necessitated allocating most samples for training, and unlike discriminative models, the testing set in a generative model serves as input for producing novel motif variations rather than as an accuracy benchmark. The seven images were distributed across three evaluation scenarios: motif and shape transformation, using Pakpak Bharat images (simple horizontal structures) and Toba images (complex diamond and zigzag patterns); colour variation (recolouring), using Toba, Simalungun, and Karo images to assess colour consistency across distinct colour profiles; and high-complexity input, using three Karo images characterized by dense, layered, and highly detailed geometric patterns.

In addition to the image dataset, a caption dataset was manually constructed by describing the visual characteristics of each Ulos motif. Each caption followed a structured template consisting of {[}Main Motif{]} + {[}Shape Details{]} + {[}Arrangement{]} + {[}Colour{]} + {[}Background{]} + {[}Cultural Context{]}, describing the primary motif, constituent geometric shapes, motif arrangement, dominant colours, background characteristics, and the corresponding cultural or sub-ethnic origin. These manually created captions were later used as the initial input to LLaVA[16][17], rather than as the representation-conditioning input.

\subsection{Proposed Model Architecture}

The proposed framework is built upon three integrated model components: Stable Diffusion XL (SDXL) v1.0 as the latent diffusion backbone, Low-Rank Adaptation (LoRA) for parameter-efficient fine-tuning, and the multimodal large language model LLaVA 1.5-7B. Figure 1 illustrates the overall architecture of the proposed framework, including the image and caption datasets, the fine-tuning process of both models, and the integration of four conditioning mechanisms text, image, representation, and semantic map into the SDXL image generation pipeline.

\includegraphics[width=4.82986in,height=2.71875in]{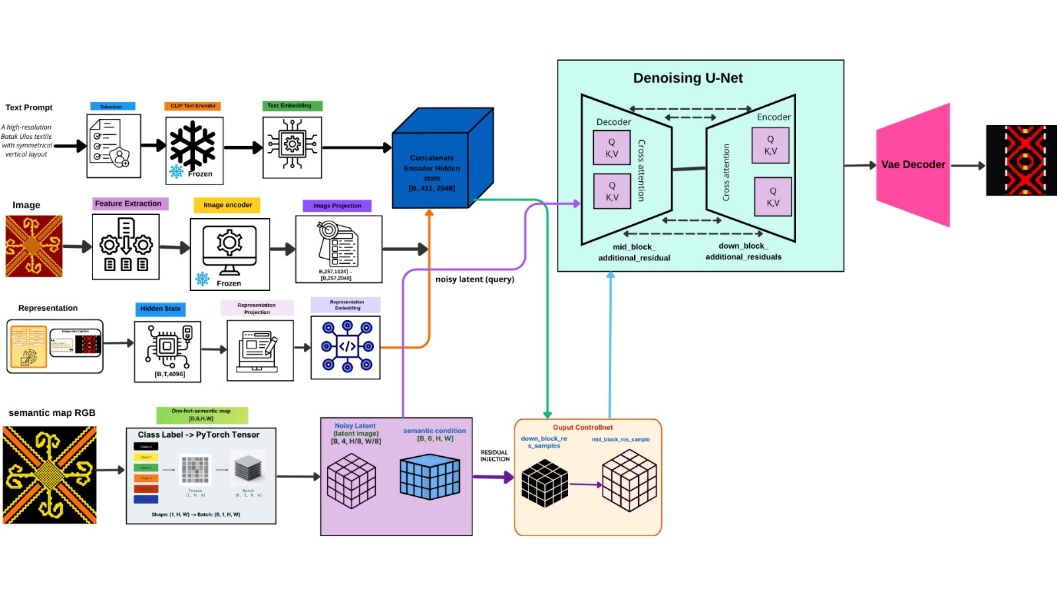}

\textbf{Fig. 1.} Proposed Generative Framework Architecture

\textbf{Backbone:} Stable Diffusion XL (SDXL) v1.0. SDXL v1.0 was selected as the primary generative model for its latent-space efficiency and greater stability against mode collapse compared to GANs. Its U-Net backbone (\textasciitilde2.6B parameters) combines two text encoders: OpenCLIP ViT-bigG and CLIP ViT-L, for richer semantic understanding of prompts, with transformer blocks configured as {[}0, 2, 10{]} across high-, medium-, and low-resolution stages to balance efficiency and semantic reasoning. A refiner model further enhances visual details before the VAE decoder reconstructs the 128×128×4 latent into a final 1024×1024×3 image.

\textbf{Fine-Tuning with LoRA}. Since SDXL and LLaVA already carry strong pretrained visual-linguistic knowledge, domain adaptation to Ulos motifs was performed using Low-Rank Adaptation (LoRA), which trains lightweight low-rank matrices within attention layers while keeping original weights frozen. LoRA was applied to all attention layers of SDXL\textquotesingle s U-Net and ControlNet, and to the Vicuna-7B language model in LLaVA (rank = 16, α = 32, dropout = 0.1), targeting the query, key, value, output, and gated MLP projections.

\textbf{Multimodal LLaVA-1.5-7B}. LLaVA generates structured visual descriptions of Ulos motifs used as representation conditioning. Its frozen CLIP ViT-L/14 vision encoder converts a 1024×1024 input image into 257×1024 patch embeddings, which a trainable two-layer MLP projects into Vicuna\textquotesingle s 4096-dimensional language space (257×4096). The LoRA-adapted Vicuna-7B model then generates captions across 32 self-attention layers, trained via cross-entropy loss against manually annotated descriptions. In conclusion, this fine-tuned adapter produces structured descriptions across seven sections: {[}Main Motif{]}, {[}Shape Details{]}, {[}Arrangement{]}, {[}Structure{]}, {[}Colour{]}, {[}Texture{]}, and {[}Cultural Context{]}, which serve as representation conditioning for the SDXL pipeline.

\subsection{Multimodal Conditioning Mechanism}

Conditioning is the primary component guiding Ulos motif generation in SDXL. This study designs a multimodal scheme combining four complementary mechanisms: text, image, representation, and semantic map conditioning, each targeting a distinct aspect of control: semantic intent, structural fidelity, visual richness, and spatial precision, respectively. This combination addresses the limitations of single-conditioning approaches (e.g., text-only or image-only), which fail to simultaneously preserve controllability and cultural authenticity.

Text conditioning provides explicit semantic control over desired motif characteristics and serves as the primary mechanism for exploring motif variations beyond the training data. It directs high-level visual attributes: motif type, base shape, pattern arrangement, and colour composition through text embeddings processed via cross-attention. Prompts are manually composed by the researcher following the Structured Captions technique, comprising four components: main subject (Ulos type and sub-ethnic origin), visual elements (geometric ornaments), compositional arrangement (symmetry and repetition), and technical-aesthetic attributes (thread texture and lighting), alongside a negative prompt excluding distorted, blurry, or non-traditional styles. An example of text conditioning is shown below

\begin{itemize}
\item
  Positive Prompt: ``A high-resolution Batak Ulos textile with symmetrical vertical layout, featuring geometric motifs such as stars, squares, rectangles, and floral patterns. The motif shapes and arrangements are creatively modified and reinterpreted while preserving the overall symmetry and traditional structure. The textile retains clear patterns, fine thread textures, and cultural visual identity.''
\item
  Negative Prompt: ``blurry, low resolution, distorted pattern, broken symmetry, messy layout, unrealistic shapes, glitch, noisy texture, cartoon style''
\end{itemize}

Image conditioning preserves the structural fidelity of the generation relative to a reference Ulos motif, preventing the process from drifting away from authentic visual structure. It acts as a visual anchor constraining geometric conventions, while the strength parameter provides explicit control over the trade-off between structural fidelity and transformation freedom. The reference image is converted by the VAE Encoder into latent space, and this latent representation, rather than the raw image, serves as the initial condition for denoising in an image-to-image generation scheme. Lower strength values preserve the motif structure more strictly, while higher values allow greater visual transformation.

Representation conditioning aims to enrich the model\textquotesingle s semantic understanding beyond the limitations of human-authored text prompts, which often cannot fully and consistently capture a motif\textquotesingle s visual details. Its contribution is to provide a more detailed and objective visual description of the reference image, while also enabling disentangled control over specific visual attributes meaning that modifying one descriptive attribute affects only the relevant visual portion without disrupting the overall motif structure. Unlike text conditioning, which originates from researcher-authored prompts, representation conditioning is a textual description automatically generated by LLaVA after analyzing the reference image, which is then projected through SDXL\textquotesingle s CLIP text encoder into a mathematical representation treated as a concept token. This approach adapts a cross-attention-based disentanglement framework (EncDiff) that theoretically exploits the time-varying information bottleneck property of the reverse diffusion process[19]. Mathematically, the representation-conditioning objective is formulated as

\begin{equation}
\mathcal{L}_{\mathrm{repr}} =
\mathbb{E}_{z_0,\epsilon,t,c}
\left[\left\|\epsilon-\epsilon_{\theta}(z_t,t,c)\right\|_2^2\right],
\label{eq:representation}
\end{equation}

where $z_0$ denotes the original image latent, $z_t$ is its noisy representation at diffusion timestep $t$, $\epsilon$ is the sampled Gaussian noise, $\epsilon_{\theta}$ denotes the noise predicted by the diffusion model, and $c$ represents the conditioning representation extracted from the LLaVA-generated description. The cross-attention mechanism uses the spatial features of the image as the Query $Q$ and the embeddings derived from the LLaVA-generated description as the Key $K$ and Value $V$. The attention operation is formulated as

\begin{equation}
\operatorname{Attention}(Q,K,V)=
\operatorname{softmax}\!\left(\frac{QK^{\mathsf T}}{\sqrt{d_k}}\right)V,
\label{eq:attention}
\end{equation}

where $d_k$ denotes the dimensionality of the key vectors. Through this mechanism, visual features attend to the most relevant textual tokens in the LLaVA-generated description, allowing specific motif attributes, such as shape, arrangement, colour, and texture, to receive targeted semantic guidance during the denoising process. Compared with ordinary text conditioning, which provides general generation instructions, representation conditioning supplies a structured and image-specific description of the reference motif. Consequently, it provides finer-grained guidance over individual visual attributes while preserving the broader structural context of the Ulos motif [19]. An example of the representation conditioning is presented below.

\begin{longtable}[]{@{}
  >{\raggedright\arraybackslash}p{(\columnwidth - 2\tabcolsep) * \real{0.3318}}
  >{\raggedright\arraybackslash}p{(\columnwidth - 2\tabcolsep) * \real{0.6682}}@{}}
\toprule\noalign{}
\endhead
\bottomrule\noalign{}
\endlastfoot
\textbf{Motif} & \textbf{Representation} \\
\includegraphics[width=1.125in,height=1.125in]{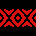} & \emph{\textbf{{[}Main Motif{]}:} The primary geometric theme of this Ulos motif is a structured 7-section visual representation. The motif consists of red shapes arranged in a repeating pattern.}

\emph{\textbf{{[}Shape Details{]}}: The motif includes stars and rectangles, which are positioned and spaced in a balanced and symmetrical manner.}

\emph{\textbf{{[}Arrangement{]}:} The shapes are organized in a repeating pattern, with each section being slightly different from the previous one.}

\emph{\textbf{{[}Structure{]}}: The structure of the motif is based on symmetry principles and a pattern flow that creates a cohesive and visually appealing design.}

\emph{\textbf{{[}Colour{]}:} The colour palette of the motif is primarily red, with variations in shades and hues. The colour is used to create a bold and striking visual effect.}

\emph{\textbf{{[}Texture{]}:} The texture of the motif is characterized by fine, intricate patterns created by the interlocking shapes. The thread characteristics and weave patterns add to the overall visual interest of the design.}

\emph{\textbf{{[}Cultural Context{]}:} This Ulos motif is a traditional Batak textile design that has been passed down through generations. It holds cultural significance and is a testament to the weaving heritage of the Batak people.} \\
\end{longtable}

Semantic map conditioning provides explicit spatial guidance over the geometric layout, decorative elements, and background of the motif, control that text or image conditioning alone cannot fully guarantee, particularly for structurally complex motifs. It prevents structural drift and layout artefacts by forcing the denoising process to respect spatial boundaries between motif elements, keeping the main motif, line structure, closed shapes, and curved ornaments consistent with Ulos conventions even when colour or texture is altered. The semantic map is constructed through a shape-based (rather than colour-based) segmentation pipeline with four stages: (1) Otsu thresholding for initial binarization of motif versus background, with the threshold computed from the pixel intensity histogram; (2) Connected Component Analysis (CCA) to group pixels into objects by size and position; (3) Hough Line Transform to detect dominant lines; and (4) Contour Analysis to distinguish closed contours (fields/frames) from high-curvature contours (curved/tendril ornaments). These stages yield six semantic classes, summarized in Table 1.

\textbf{Table 1.} Semantic Map Class Encoding Scheme for ControlNet Conditioning

\begin{longtable}[]{@{}
  >{\raggedright\arraybackslash}p{(\columnwidth - 6\tabcolsep) * \real{0.0962}}
  >{\raggedright\arraybackslash}p{(\columnwidth - 6\tabcolsep) * \real{0.2115}}
  >{\raggedright\arraybackslash}p{(\columnwidth - 6\tabcolsep) * \real{0.2500}}
  >{\raggedright\arraybackslash}p{(\columnwidth - 6\tabcolsep) * \real{0.4423}}@{}}
\toprule\noalign{}
\endhead
\bottomrule\noalign{}
\endlastfoot
Class & Region Name & Reference RGB Clour & Geometric Indicator \\
0 & Background & (0, 0, 0) -- black & Area outside the Otsu thresholding result \\
1 & Main Motif & (255, 215, 0) -- yellow & Largest/dominant connected component \\
2 & Line Structure & (50, 205, 50) -- green & Dominant straight lines from the Hough Line Transform \\
3 & Closed field/frame & (255, 140, 0) -- orange & Closed contours from Contour Analysis \\
3 & Curved ornament & (220, 20, 60) -- red & Non-linear contours (high curvature \\
\end{longtable}

In the SDXL-ControlNet pipeline, RGB pixels of the semantic map are converted into class indices (Class ID 0--5) via nearest-neighbor matching to six reference colours rather than thresholding, to avoid errors from compression and anti-aliasing. The class-ID map is then one-hot encoded into a (B, 6, H, W) tensor at full image resolution, without downsampling to latent resolution. Rather than concatenating directly into the UNet\textquotesingle s input channels, ControlNet is initialized as a copy of the UNet structure with six conditioning channels, trained independently while the main SDXL UNet remains frozen at its standard 4 channels. ControlNet receives the noisy latent (B, 4, H/8, W/8) together with the controlnet\_cond, producing residuals from each down block and mid-block that are injected into the SDXL UNet via down\_block\_additional\_residuals and mid\_block\_additional\_residual. Thus, the semantic map influences denoising purely through residual injection from the trained ControlNet, rather than by modifying the UNet\textquotesingle s input channels. Fig. 2 shows an example of semantic map conditioning implementation.

\includegraphics[width=3.14236in,height=3.19722in]{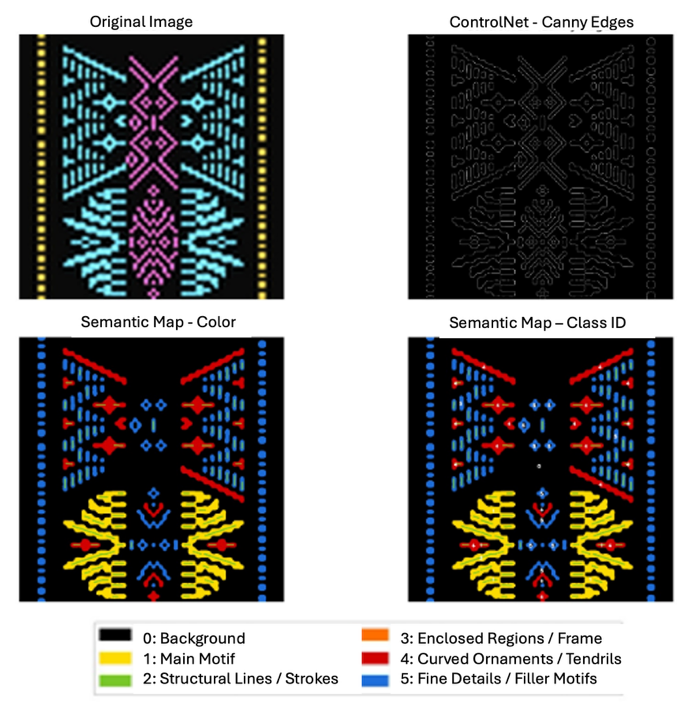}

\textbf{Fig. 2.} Example of Semantic Map Implementation Results

To evaluate the effectiveness of the proposed conditioning framework, a five-level ablation study was conducted by progressively incorporating additional conditioning modalities into the SDXL generation pipeline. The evaluated configurations were: (1) the baseline SDXL model without fine-tuning using only text and image conditioning, (2) the LoRA fine-tuned SDXL model with text and image conditioning, (3) the addition of semantic map conditioning through ControlNet, (4) the incorporation of representation conditioning generated by LLaVA, and (5) the complete framework integrating text, image, representation, and semantic map conditioning. Each configuration was evaluated using the same testing images under identical experimental settings.

To comprehensively assess the controllability and robustness of the proposed framework, three evaluation scenarios were designed. Each scenario employed customized positive and negative prompts to guide the generation process according to specific experimental objectives.

\section{Result and Discussion}
\subsection{Training Result}

\textbf{Training LLaVA.} The LLaVA model was refined to enable it to understand and recognize geometric patterns, distinctive colours, and the structure of Ulos motifs, enabling it to produce specific and accurate descriptions of Ulos motifs. Refinement was conducted over 20 epochs using 431 pairs of Ulos motifs, each accompanied by a brief description of the motif.

\begin{longtable}[]{@{}
  >{\raggedright\arraybackslash}p{(\columnwidth - 2\tabcolsep) * \real{0.5021}}
  >{\raggedright\arraybackslash}p{(\columnwidth - 2\tabcolsep) * \real{0.4979}}@{}}
\toprule\noalign{}
\endhead
\bottomrule\noalign{}
\endlastfoot
\begin{minipage}[t]{\linewidth}\raggedright
\includegraphics[width=2.32708in,height=1.41597in]{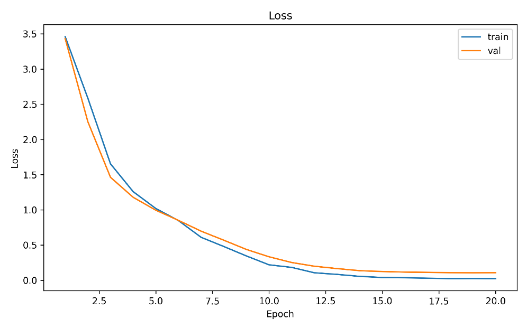}

\begin{enumerate}
\def\labelenumi{(\alph{enumi})}
\tightlist
\item
\end{enumerate}
\end{minipage} & \begin{minipage}[t]{\linewidth}\raggedright
\includegraphics[width=2.30764in,height=1.41944in]{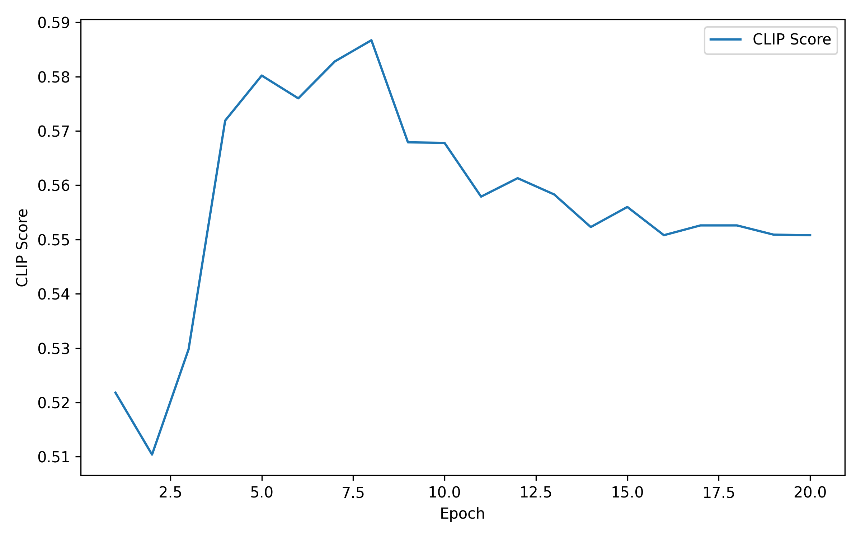}

\begin{enumerate}
\def\labelenumi{(\alph{enumi})}
\setcounter{enumi}{1}
\tightlist
\item
\end{enumerate}
\end{minipage} \\
\multicolumn{2}{@{}>{\raggedright\arraybackslash}p{(\columnwidth - 2\tabcolsep) * \real{1.0000} + 2\tabcolsep}@{}}{%
\begin{minipage}[t]{\linewidth}\raggedright
\includegraphics[width=2.41875in,height=1.45556in]{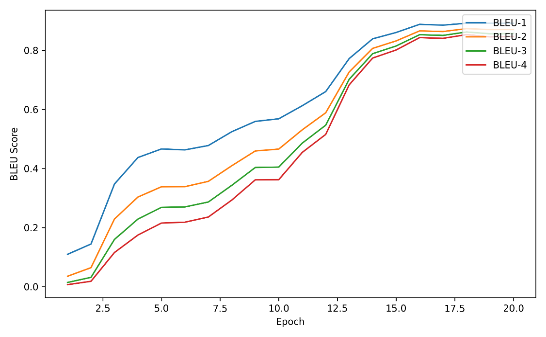}

\begin{enumerate}
\def\labelenumi{(\alph{enumi})}
\setcounter{enumi}{2}
\tightlist
\item
\end{enumerate}
\end{minipage}} \\
\end{longtable}

\textbf{Fig. 3.} Training performance of the fine-tuned LLaVA model over 20 epochs: (a) training and validation loss, (b) CLIP Score, and (c) BLEU-1 to BLEU-4 scores.

As shown in the loss curve (a), both training and validation loss decrease sharply during the first five epochs from approximately 3.5 to around 1.0, indicating that the model rapidly learns the dominant structural and colour patterns of the Ulos motifs early in training. Beyond epoch 5, the loss continues to decline at a more gradual rate, eventually converging to near-zero values (below 0.1) by epoch 20. Throughout training, the training and validation curves remain closely aligned, with no divergence or widening gap between them at any point. This close correspondence indicates that the fine-tuned model generalizes well to unseen validation samples rather than merely memorizing the training data, confirming that overfitting did not occur despite the relatively small dataset size. The smooth, monotonic convergence of both curves further suggests that the learning rate and LoRA rank configuration were well-suited for adapting SDXL to the Ulos motif domain without unstable or oscillatory training behaviour.

CLIPScore measures the alignment between LLaVA-generated descriptions and their corresponding captions. In the CLIPScore graph (b), at the end of training, the CLIPScore value is in the range of 0.55--0.56. The final CLIPScore value during training indicates that LLaVA successfully learned and understood the characteristics of the Ulos motif.

BLEU is used to measure how closely the descriptions generated by LLaVA match the researchers\textquotesingle{} manual captions. In the BLEU graph (c), the BLEU score (1--4) improves during training, approaching a value close to 1 (±0.90) in the final epoch. This indicates that the model can produce descriptions that closely match the reference captions.

\textbf{Training SDXL.} The SDXL model was trained to recognize the geometric patterns of Ulos motifs and to generate Ulos motifs that remain consistent with the characteristics of the Ulos motif. Training was conducted through four SDXL training experiments of 50 epochs each, with each training session using a different conditioning combination. Table 2 below summarizes the training results for each conditioning combination.

\textbf{Table 2.} Quantitative Comparison of SDXL Fine-Tuning Performance Across Conditioning Configurations (E1--E4)

\begin{longtable}[]{@{}
  >{\raggedright\arraybackslash}p{(\columnwidth - 8\tabcolsep) * \real{0.3200}}
  >{\raggedright\arraybackslash}p{(\columnwidth - 8\tabcolsep) * \real{0.1600}}
  >{\raggedright\arraybackslash}p{(\columnwidth - 8\tabcolsep) * \real{0.1401}}
  >{\raggedright\arraybackslash}p{(\columnwidth - 8\tabcolsep) * \real{0.1600}}
  >{\raggedright\arraybackslash}p{(\columnwidth - 8\tabcolsep) * \real{0.2199}}@{}}
\toprule\noalign{}
\endhead
\bottomrule\noalign{}
\endlastfoot
Experiment (Conditioning) & Loss & Best FID & Stable SSIM & CLIP Score Stable \\
E1 (blue): Text, Image & ±0,003 & ±303 & 0,84 & 0,50--0,60 \\
E2 (green): Text, Image, Semantic Map & ±0,003 & ±270 & 0,65 & 0,65--0,70 \\
E3 (yellow): Text, Image, Representation & 0,0024 & ±280 & 0,84 & 0,60--0,65 \\
E4 (red): Text, Image, Semantic Map, Representation & ±0,0025 & ±330 & 0,84 & 0,55--0,60 \\
\end{longtable}

Fig. 4 summarizes the training dynamics and generation quality of SDXL across the four conditioning configurations over 50 epochs. All configurations show a similar pattern of rapid loss reduction within the first 10-15 epochs before stabilizing, indicating that each variant successfully converges during fine-tuning (a). However, clear differences emerge in validation loss (b): the Text+Image+Semantic Map (green) and Text+Image+Representation (orange) configurations reach the lowest and most stable validation loss (\textasciitilde0.005--0.006), while Text+Image (blue) plateaus at a noticeably higher level (\textasciitilde0.010), and the full four-conditioning configuration (red) settles at an intermediate value (\textasciitilde0.007) with mild fluctuation in later epochs. This suggests that adding Representation or Semantic Map conditioning individually improves generalization beyond text-and-image conditioning alone, whereas combining all four signals simultaneously introduces some optimization difficulty.

In terms of generation quality, a trade-off emerges between distributional fidelity and structural consistency. The Semantic Map configuration (green) and Representation configuration (orange) achieve the lowest and most stable FID scores (c), indicating outputs statistically closer to real Ulos motifs, and the Semantic Map configuration also attains the highest and most consistent CLIP Score (e), reflecting stronger semantic alignment with textual descriptions. However, the Semantic Map configuration shows lower SSIM (d) values and greater volatility, suggesting reduced structural stability relative to the reference motifs. In contrast, the Text+Image+Representation (orange) configuration maintains consistently high SSIM (\textasciitilde0.83) alongside low, stable validation loss and competitive FID, indicating the best overall balance between structural fidelity, distributional quality, and training stability. The full four-conditioning configuration (red), despite combining all conditioning signals, exhibits the weakest and most unstable FID and CLIP Scores, suggesting that jointly enforcing Representation and Semantic Map constraints may introduce conflicting optimisation signals rather than purely additive benefits. Overall, the Text, Image, Representation configuration demonstrates the most consistently good and stable learning behaviour across all five metrics.

\includegraphics[width=4.84931in,height=2.47708in]{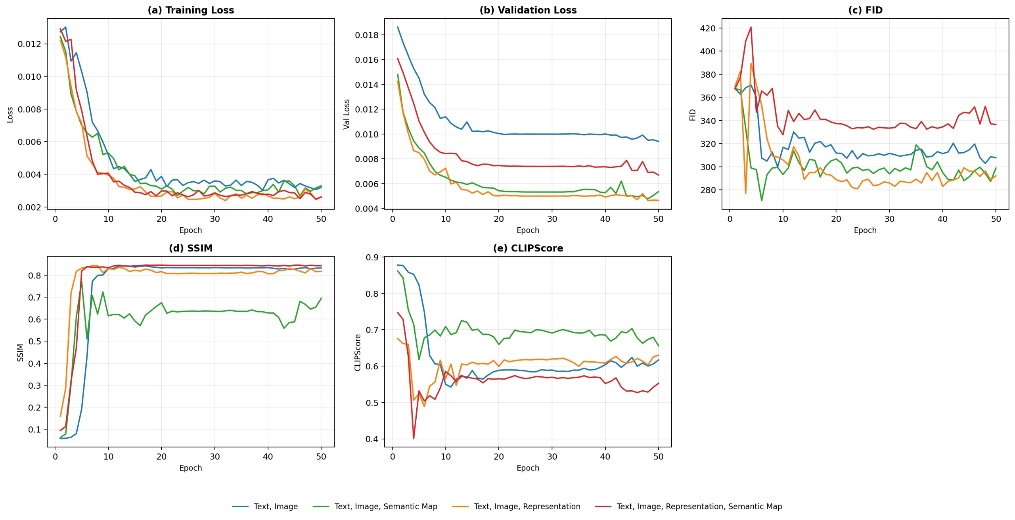}

\textbf{Fig. 4.} SDXL Fine-Tuning Performance Across Conditioning Configurations

\subsection{Testing Result}

Model testing was conducted across three scenarios: (1) changes in shape and motif, (2) colour variation/recolouring, and (3) high-complexity input. For each scenario, generated outputs were compared against the baseline (SDXL without fine-tuning) and the four experimental configurations (E1--E4), with results selected based on the best-performing FID and SSIM scores, alongside their corresponding reference conditioning inputs.

Table 3 presents a qualitative comparison of the source motifs and their corresponding generated outputs across the three testing scenarios, with each row representing the best result selected according to FID, SSIM, and CLIP Score, respectively. In Scenario 1 (shape/motif transformation), all configurations successfully preserve the core X-diamond structure of the source motif, with visible improvement in edge sharpness and pattern regularity from the baseline through E1--E4, indicating that additional conditioning signals progressively refine structural clarity without altering the fundamental motif shape. In Scenario 2 (colour variation), the baseline and early experiments show noticeable colour instability and structural distortion relative to the reference green-and-orange motif, whereas E3 and E4 exhibit more consistent colour reproduction and cleaner pattern boundaries, suggesting that Representation and Semantic Map conditioning play a key role in stabilising colour fidelity during recolouring tasks. In Scenario 3 (high-complexity input), all configurations manage to reproduce the dominant diamond-lattice structure of the dense reference motif, though subtle differences emerge in the preservation of fine internal cross-hatching and border details, with later experiments (E3, E4) showing marginally sharper detail retention in the most visually intricate regions. Overall, the results indicate that as task difficulty increases from simple shape transformation to complex recolouring and high-density motifs, the benefit of combining multiple conditioning mechanisms becomes more apparent, particularly in maintaining both structural integrity and colour/detail accuracy.

\textbf{Table 3.} Visual Comparison of Source and Generated Motifs Across Testing Scenarios 1--3 Under Baseline and Ablation Configurations (E1--E4)

\begin{longtable}[]{@{}
  >{\raggedright\arraybackslash}p{(\columnwidth - 4\tabcolsep) * \real{0.1812}}
  >{\raggedright\arraybackslash}p{(\columnwidth - 4\tabcolsep) * \real{0.1995}}
  >{\raggedright\arraybackslash}p{(\columnwidth - 4\tabcolsep) * \real{0.6193}}@{}}
\toprule\noalign{}
\endhead
\bottomrule\noalign{}
\endlastfoot
Scenario & Source & Motif Generated \\
1 & \includegraphics[width=0.62222in,height=0.61736in]{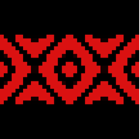} & \includegraphics[width=2.60417in,height=1.57014in]{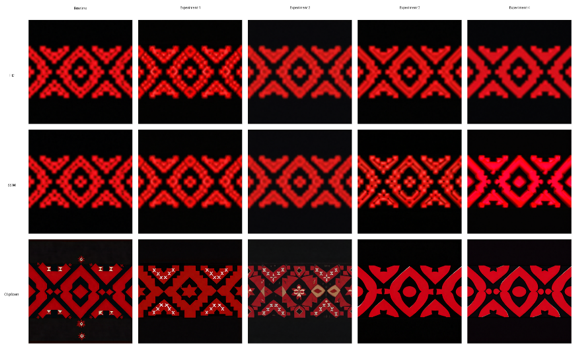} \\
2 & \includegraphics[width=0.61597in,height=0.61597in]{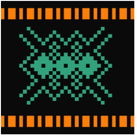} & \includegraphics[width=2.59861in,height=1.57361in]{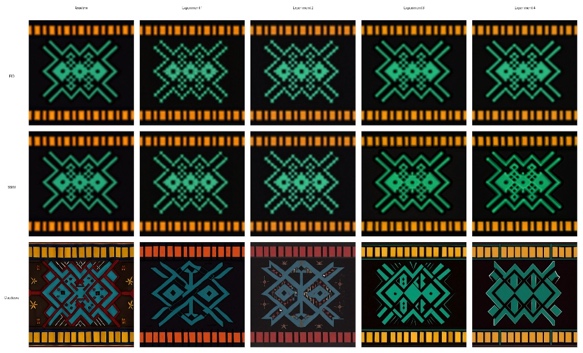} \\
3 & \includegraphics[width=0.62986in,height=0.62431in]{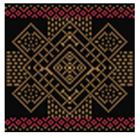} & \includegraphics[width=2.60347in,height=1.57083in]{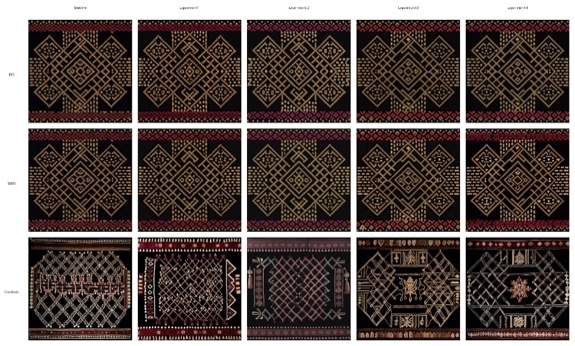} \\
\end{longtable}

\subsection{Evaluation}

The evaluation was conducted using two methods: quantitative and qualitative. Quantitative evaluation was conducted to assess the quality and diversity of Ulos motifs generated by the Stable Diffusion model. A qualitative evaluation was conducted to assess the generated Ulos motifs based on visual aesthetics and cultural values.

\textbf{Quantitative Evaluation}

A quantitative evaluation was conducted to systematically assess the performance of each experimental configuration across varying generation settings, specifically the number of inference steps, the strength parameter (controlling the degree of deviation from the reference image), and the guidance scale (controlling adherence to the text prompt). These parameters were varied across all experiments to examine their influence on the quality, fidelity, and controllability of the generated Ulos motifs, using FID, SSIM, and CLIP Score as the primary evaluation metrics.

Among the evaluated diffusion steps (30, 40, and 50), 30 steps achieved the best overall performance. It produced the highest CLIP Image Score (0.9122), CLIP Consistency Score (0.9235), Structural Score (0.8418), and SSIM (0.7103). Increasing the number of diffusion steps beyond 30 did not improve semantic alignment and instead reduced structural fidelity, indicating that excessive denoising tends to alter fine Ulos motif characteristics. Therefore, 30 diffusion steps were selected as the optimal configuration for subsequent experiments.

Based on the Fréchet Inception Distance (FID), 30 diffusion steps achieved the best overall image quality, yielding the lowest average FID score (253.67) across all experimental scenarios. Increasing the diffusion process to 40 and 50 steps resulted in progressively higher FID values (279.00 and 289.38, respectively), indicating that additional denoising did not improve realism and instead caused the generated motifs to deviate further from the distribution of authentic Ulos patterns. Although one configuration in Scenario 1 achieved its minimum FID at 50 steps, the overall experimental results demonstrate that 30 steps provide the most consistent and realistic image generation performance.

The SSIM analysis indicates that 30 diffusion steps achieve the highest structural fidelity, with an average SSIM of 0.7103. Increasing the number of diffusion steps to 40 and 50 reduces the structural similarity to the reference motifs, yielding average SSIM values of 0.6775 and 0.6831, respectively. These findings suggest that excessive denoising progressively modifies the characteristic geometry and motif arrangement of Ulos textiles, whereas 30 steps provide the best balance between image synthesis and preservation of the original cultural patterns.

Fig. 5 presents the evaluation results for Scenario 2 (colour variation/recolouring) using the best-performing configuration identified during testing, generated with a fixed number of inference steps (steps = 30), across varying strength and guidance scale values.

\includegraphics[width=4.53333in,height=3.87847in]{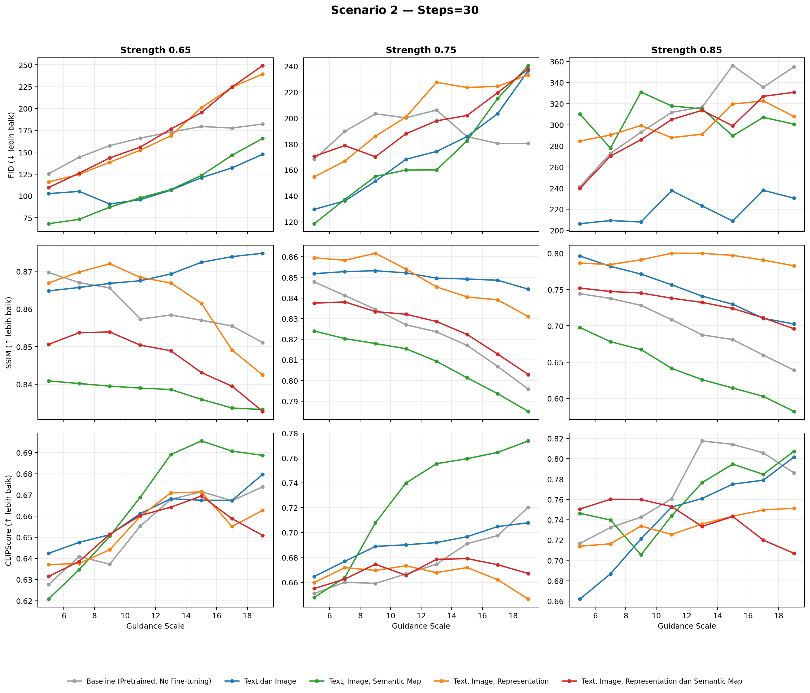}

\textbf{Fig. 5.} Effect of Guidance Scale and Strength on FID, SSIM, and CLIP Score for the Baseline and Four Conditioning Configurations in Scenario 2 (Colour Variation/Recolouring) at 30 Inference Steps

\textbf{Qualitative Evaluation.}

Qualitative evaluation was conducted to capture human perceptions of the generated Ulos motifs through questionnaires distributed to two respondent groups: traditional weavers and the public participants. Weavers (n = 9, from Desa Pintu Bantu, with 2--40 years of weaving experience) assessed the motifs based on cultural and structural fidelity, pattern authenticity, colour alignment with tradition, and structural feasibility of the woven texture. Across the three scenarios, Scenario 1 achieved moderate scores (3.5--4.0) for pattern authenticity and structural feasibility but slightly lower colour alignment, Scenario 2 showed improved structural feasibility while maintaining stable authenticity and colour scores, and Scenario 3 achieved the highest ratings overall, approaching 4.5 for both pattern authenticity and structural feasibility. Weavers also perceived the generated motifs as sufficiently diverse across all scenarios, suggesting the model successfully produces varied yet recognizable Ulos patterns.

Public participants (n = 30, predominantly familiar to very familiar with Ulos culture) evaluated the motifs from an aesthetic and cultural-identity perspective, covering visual composition, modern/innovative impression, and alignment with distinctive Ulos characteristics. All average scores across the three scenarios exceeded the Likert midpoint (3.0), with Scenario 3 showing the most stable and consistently high ratings across all three aspects.

The reliability of both questionnaire instruments was confirmed by a Cronbach\textquotesingle s Alpha of 0.991 for both groups, indicating excellent internal consistency well above the accepted threshold (α \textgreater{} 0.9). Since the Shapiro-Wilk test indicated that most assessment aspects were not normally distributed (p \textless{} 0.05), a nonparametric Wilcoxon Signed-Rank Test was used to evaluate acceptance relative to the neutral Likert midpoint (test value = 3). Results showed that all aspects, across both weaver and public groups and all three scenarios, yielded Asymp. Sig. (2-tailed) values below 0.05, indicating that both respondent groups rated the model-generated Ulos motifs as statistically significantly above the neutral midpoint, providing strong evidence that the generated motifs were positively received in terms of both cultural authenticity and aesthetic quality.

\section{Conclusion}

This study successfully designed and implemented a multimodal generative framework that integrates a Latent Diffusion Model (Stable Diffusion XL v1.0) with a Multimodal Large Language Model (LLaVA 1.5-7B) through four complementary conditioning mechanisms: text, image, representation, and semantic map, to generate diverse Batak Ulos motifs that remain both controllable and faithful to their cultural characteristics. The training results reveal that conditioning effectiveness is not strictly proportional to the number of conditioning signals combined: the Text, Image, and Semantic Map configuration achieved the best FID (±270) and the highest CLIP Score (0.65--0.70), as spatial guidance from the semantic map helped the model capture motif layout and semantic alignment more precisely; however, this same configuration also yielded the lowest SSIM (±0.65), indicating reduced structural stability. In contrast, the Text, Image, and Representation configuration maintained consistently high SSIM (±0.84) alongside a competitive FID (±280), representing the best overall balance among the four configurations. Combining all four conditioning mechanisms simultaneously produced the weakest FID (±330), suggesting that jointly enforcing multiple conditioning constraints introduces competing optimization signals rather than purely additive benefits. During testing, increasing the guidance scale consistently improved both FID and CLIP Score across all scenarios, whereas SSIM showed less consistent performance as the generated motifs increasingly diverged structurally from their reference images.

Qualitative evaluation through questionnaires administered to traditional Ulos weavers (n = 9) and public participants (n = 30) demonstrated statistically significant and positive acceptance of the generated motifs (Wilcoxon Signed-Rank Test, p = 0.007 for weavers and p \textless{} 0.001 for public participants across all assessed aspects). Notably, Scenario 3 (high-complexity input) achieved the highest and most stable scores across both respondent groups, indicating that the generated motifs were well received for both cultural authenticity and visual aesthetics, even under the most structurally demanding testing conditions.

Overall, these findings confirm that combining multiple, complementary conditioning mechanisms, rather than simply maximising their number, offers an effective strategy for balancing generation quality, structural fidelity, and cultural authenticity in AI-assisted heritage textile design. For future work, several directions merit exploration: expanding the Ulos motif dataset, particularly for sub-ethnic groups with more complex patterns; broadening the qualitative evaluator pool to include art academics, museum curators, and Batak culture enthusiasts; further developing the web prototype, for instance through a generation-history feature; and investigating the inclusion of "stick motif" as an additional conditioning modality to further enrich the framework\textquotesingle s motif generation capability.

\section*{References}

1. Hasibuan, R.A., Rochmat, S.: Ulos as Batak Cultural Wisdom Towards World Heritage. birle, budapest. internation. research. and. critic. in. linguistic. educatie. 4, 853--864 (2021). https://doi.org/10.33258/birle.v4i2.1865.

2. Nugroho, C., Nurhayati, I.K., Nasionalita, K., Malau, R.M.U.: Weaving and Cultural Identity of Batak Toba Women. Journal of Asian and African Studies. 56, 1165--1177 (2021). https://doi.org/10.1177/0021909620958032.

3. Siagian, R.J.: The Symbolic Meaning of Traditional Woven Fabric Ulos as A Spiritual Expression in Batak Toba Rituals. Intl. J. Rel. 5, 200--209 (2024). https://doi.org/10.61707/stw03g83.

4. Erlyana, Y.: Kajian Visual Keragaman Corak Pada Kain Ulos. JDD. 1, 35--46 (2016). https://doi.org/10.25105/jdd.v1i1.408.

5. Adriani, A., Fitriani, N.: Motif dan Makna Motif Tenun Ulos Batak Angkola di Kabupaten Tapanuli Selatan. GR. 12, 302--309 (2023). https://doi.org/10.24114/gr.v12i2.49593.

6. Aruan, A.D.K.: Etnomatematika: Eksplorasi Motif Ulos Batak Toba Dalam Pembelajaran Bangun Datar. sepren. 5, 57--65 (2024). https://doi.org/10.36655/sepren.v5i02.1403.

7. Azzahra, I.R., Siregar, N.A., Karosekali, Y.M., Salim, Z.F., Siregar, T.M.S.B.: The Meaning of Mangolusi in The Batak Toba Wedding Ritual. Multidiciplinary Scientifict~Journal. 2, 263--269 (2024). https://doi.org/10.57185/mutiara.v2i5.179.

8. Raad, L., Galerne, B.: Efros and Freeman Image Quilting Algorithm for Texture Synthesis. Image Processing OnLine. 7, 1--22 (2017). https://doi.org/10.5201/ipol.2017.171.

9. Shaham, T.R., Dekel, T., Michaeli, T.: SinGAN: Learning a Generative Model from a Single Natural Image, https://arxiv.org/abs/1905.01164, (2019). https://doi.org/10.48550/ARXIV.1905.01164.

10. Simanjuntak, H., Panjaitan, E., Siregar, S., Manalu, U., Situmeang, S., Barus, A.: Generating New Ulos Motif with Generative AI Method in Digital Tenun Nusantara (DiTenun) Platform. ijacsa. 15, (2024). https://doi.org/10.14569/IJACSA.2024.01507109.

11. Simanjuntak, H.T.A., Purba, J., Girsang, S., Manurung, W., Situmeang, S., Barus, A., Siahaan, D.O.: AI for Cultural Heritage Textiles: Fine-Tuned Latent Diffusion for Novel Ulos Motif Synthesis, https://arxiv.org/abs/2607.06590, (2026). https://doi.org/10.48550/ARXIV.2607.06590.

12. Octadion, O., Yudistira, N., Kurnianingtyas, D.: Synthesis of batik motifs using a diffusion - generative adversarial network. Multimed Tools Appl. 84, 3407--3438 (2025). https://doi.org/10.1007/s11042-025-20620-9.

13. Dhariwal, P., Nichol, A.: Diffusion Models Beat GANs on Image Synthesis, https://arxiv.org/abs/2105.05233, (2021). https://doi.org/10.48550/ARXIV.2105.05233.

14. Rombach, R., Blattmann, A., Lorenz, D., Esser, P., Ommer, B.: High-Resolution Image Synthesis with Latent Diffusion Models. In: 2022 IEEE/CVF Conference on Computer Vision and Pattern Recognition (CVPR). pp. 10674--10685. IEEE, New Orleans, LA, USA (2022). https://doi.org/10.1109/CVPR52688.2022.01042.

15. Podell, D., English, Z., Lacey, K., Blattmann, A., Dockhorn, T., Müller, J., Penna, J., Rombach, R.: SDXL: Improving Latent Diffusion Models for High-Resolution Image Synthesis, https://arxiv.org/abs/2307.01952, (2023). https://doi.org/10.48550/ARXIV.2307.01952.

16. Liu, H., Li, C., Wu, Q., Lee, Y.J.: Visual Instruction Tuning. In: Advances in Neural Information Processing Systems 36. pp. 34892--34916. Neural Information Processing Systems Foundation, Inc. (NeurIPS), New Orleans, Louisiana, USA (2023). https://doi.org/10.52202/075280-1516.

17. Liu, H., Li, C., Li, Y., Lee, Y.J.: Improved Baselines with Visual Instruction Tuning. In: 2024 IEEE/CVF Conference on Computer Vision and Pattern Recognition (CVPR). pp. 26286--26296. IEEE, Seattle, WA, USA (2024). https://doi.org/10.1109/CVPR52733.2024.02484.

18. Zhang, L., Rao, A., Agrawala, M.: Adding Conditional Control to Text-to-Image Diffusion Models. In: 2023 IEEE/CVF International Conference on Computer Vision (ICCV). pp. 3813--3824. IEEE, Paris, France (2023). https://doi.org/10.1109/ICCV51070.2023.00355.

19. Yang, T., Lan, C., Lu, Y., Zheng, N.: Diffusion Model with Cross Attention as an Inductive Bias for Disentanglement. In: Advances in Neural Information Processing Systems 37. pp. 82465--82492. Neural Information Processing Systems Foundation, Inc. (NeurIPS), Vancouver, BC, Canada (2024). https://doi.org/10.52202/079017-2622.

\end{document}